\documentclass[conference]{IEEEtran}
\usepackage{cite}
\usepackage{amsmath,amssymb,amsfonts}
\usepackage{algorithmic}
\usepackage{graphicx}
\usepackage{textcomp}
\usepackage{xcolor}

\usepackage{balance}
\usepackage{mathtools}
\usepackage{tikz}
\usepackage{tkz-euclide}
\usepackage{pgfplots} 
\pgfplotsset{compat=1.9}
\usetikzlibrary{fit,matrix,chains,positioning,decorations.pathreplacing,arrows}
\usepackage[mode=buildnew]{standalone}  

\newcommand{\bma}{\begin{bmatrix}}
\newcommand{\ema}{\end{bmatrix}}

\newcommand{\T}{{\mathsf{T}}} 

\newcommand{\Reals}{\mathbb{R}}      

\newcommand{\Normal}[1]{\mathcal{N}\!\left({#1}\right)} 

\newcommand{\eqdef}{\triangleq} 

\newcommand{\argmin}{\operatorname*{argmin}}
\newcommand{\argmax}{\operatorname*{argmax}}

\newcommand{\Exp}[1]{\exp \left({#1}\right)} 

\makeatletter
\DeclareFontFamily{U}{MnSymbolA}{}
\DeclareSymbolFont{MnSyA}{U}{MnSymbolA}{m}{n}
\DeclareFontShape{U}{MnSymbolA}{m}{n}{
<-6> MnSymbolA5
<6-7> MnSymbolA6
<7-8> MnSymbolA7
<8-9> MnSymbolA8
<9-10> MnSymbolA9
<10-12> MnSymbolA10
<12-> MnSymbolA12}{}
\DeclareMathSymbol{\smallrightarrow}{\mathrel}{MnSyA}{0}
\DeclareMathSymbol{\smallleftarrow}{\mathrel}{MnSyA}{2}
\DeclareMathSymbol{\smallleftrightarrow}{\mathrel}{MnSyA}{16}
\newcommand{\smallrightarrowfill@}{\arrowfill@\relbar\relbar\smallrightarrow}
\newcommand{\smallleftarrowfill@}{\arrowfill@\smallleftarrow\relbar\relbar}
\newcommand{\smallleftrightarrowfill@}
{\arrowfill@\smallleftarrow\relbar\smallrightarrow}
\renewcommand{\overrightarrow}{\mathpalette{\overarrow@\smallrightarrowfill@}}
\renewcommand{\overleftarrow}{\mathpalette{\overarrow@\smallleftarrowfill@}}
\renewcommand{\overleftrightarrow}
{\mathpalette{\overarrow@\smallleftrightarrowfill@}}
\makeatother

\newcommand{\cond}{\hspace{0.02em}|\hspace{0.08em}}

\newcounter{examplecntr}
{\begin{trivlist}\small\item[]\refstepcounter{examplecntr}%
 {\bfseries Example~\theexamplecntr%
  \ifthenelse{\equal{#1}{}}{}{ (#1)}.
}}%
{\end{trivlist}}

\newcounter{definitioncntr}
{\begin{trivlist}\item[]\refstepcounter{definitioncntr}%
{\bfseries Definition~\thedefinitioncntr.}}%
{\hfill$\Box$\end{trivlist}}

\newcounter{theoremcntr}
{\begin{trivlist}\item[]\refstepcounter{theoremcntr}%
{\bfseries Theorem~\thetheoremcntr%
  \ifthenelse{\equal{#1}{}}{}{ (#1)}.
}}%
{\hfill$\Box$\end{trivlist}}

\newcounter{propositioncntr}
{\begin{trivlist}\item[]\refstepcounter{propositioncntr}%
{\bfseries Proposition~\thepropositioncntr%
  \ifthenelse{\equal{#1}{}}{}{ (#1)}.
}}%
{\hfill$\Box$\end{trivlist}}

\newcounter{saveequationcntr}

\def\va{\sigma_a^2}
\def\vb{\sigma_b^2}
\def\vah{\hat \sigma_a^2}
\def\vbh{\hat \sigma_b^2}
\def\mt{m_\theta}
\def\vt{\sigma_\theta^2}
\def\mtt{\tilde{m}_\theta}
\def\vtt{\tilde{\sigma}_\theta^2}
\def\limb{\lim_{b \rightarrow \infty}}
\def\binf{{b \rightarrow \infty}}

\begin{document}
{}
\title{Half-Space and Box Constraints as NUV Priors: First Results}
\author{\IEEEauthorblockN{Raphael Keusch and Hans-Andrea Loeliger}
\IEEEauthorblockA{
\textit{ETH Zurich, Dept. of Information Technology \& Electrical Engineering}\\
\{keusch, loeliger\}@isi.ee.ethz.ch}
}

\maketitle

\begin{abstract}
Normals with unknown variance (NUV) can represent many useful priors
and blend well with Gaussian models and message passing algorithms.
NUV representations of sparsifying priors have long been known,
and NUV representations of binary (and $M$-level) priors
have been proposed very recently. In this document,
we propose NUV representations of half-space constraints and
box constraints,
which allows to add such constraints to any linear Gaussian model
with any of the previously known NUV priors without affecting
the computational tractability.
\end{abstract}

\section{Introduction}
NUV priors (normals with unknown variance)
hugely extend the expressive power of linear Gaussian models
while essentially maintaining their computational tractability by standard algorithms. 
NUV priors originated in sparse Bayesian 
learning~\cite{mackay1992bayesian,tipping_sparse_2001,
tipping_fast_2003,wipf_sparse_2004,wipf_new_2008,loeliger_sparsity_2016} 
and are closely related to variational representations of sparsifying 
priors~\cite{bach_optimization_2012,loeliger_factor_2018}.
However, NUV priors can also represent smoothed versions of 
such priors (including the Huber function)~\cite{loeliger_factor_2018}, 
and it has very recently been shown that 
NUV priors can also represent discretizing 
priors~\cite{keusch2021binaryNUV,keusch2021binaryNUVext}.

In this document, we show that NUV priors can also express half-space 
constraints (inequality constraints) and
box constraints (interval constraints). 
This is not difficult (with hindsight), but it is very useful,
as it allows to include such constraints in linear Gaussian 
models (with or without additional NUV priors),
while maintaining their tractability by standard algorithms such as
iterated least-squares~\cite{byrd1979convergence,schroeder1991lp} or iterated versions of Kalman-type algorithms for linear Gaussian models, 
cf.\ \cite{loeliger_sparsity_2016,loeliger_factor_2018}.

In formal terms, a half-space constraint on a quantity $x \in \Reals$ enforces
the inequality
\begin{IEEEeqnarray}{rCl}
  x \leq a \quad \text{or} \quad x \geq a, \label{eqn:HalfSpaceCon}
\end{IEEEeqnarray}
with $a \in \Reals$.
%
Similarly, a box constraint on $x$ enforces the inequality
\begin{IEEEeqnarray}{rCl}
  a \leq x \leq b, \label{eqn:BoxCon}
\end{IEEEeqnarray}
with $a, b \in \Reals$.

In the following, we will derive NUV prior representations of the 
constraints~(\ref{eqn:HalfSpaceCon}) and~(\ref{eqn:BoxCon}), 
where we assume that $x$ is a variable in some linear Gaussian
model. 

\section{NUV Representation of the Laplace Prior} 
\label{sec:LaplacePrior}
To begin with, we provide a quick primer on the 
(well-known \cite{bach_optimization_2012,loeliger_factor_2018}) NUV
representation of the Laplace prior, as this will 
be a fundamental step for the derivation of the proposed prior models. 
The Laplace prior can be represented using a NUV prior of the 
form~\footnote{We denote a Gaussian probability density in $x$ with mean $\mu$ and variance $\sigma^2$ by $\Normal{x; \mu, \sigma^2}$.}
\begin{IEEEeqnarray}{rCl}
   p(x) \eqdef \max_{\sigma^2} \Normal{x; a, \sigma^2} \rho(\sigma), 
   \label{eqn:LaplacePrior}
 \end{IEEEeqnarray} 
 where $\sigma^2$ is an unknown variance, $a \in \Reals$, and 
 \begin{IEEEeqnarray}{rCl}
   \rho(\sigma) \eqdef \sqrt{2 \pi \sigma^2} \Exp{- \gamma^2 \sigma^2 / 2},
 \end{IEEEeqnarray}
with $\gamma > 0$. Note that for fixed 
$\sigma^2$,~(\ref{eqn:LaplacePrior})
is Gaussian, up to a scale factor.

Such variational representations of non-Gaussian
priors in combination with an otherwise Gaussian model blend well with iterative
algorithms that alternate between estimating $x$ (for fixed $\sigma^2$), for
instance by least-squares or Kalman-type algorithms, 
and estimating $\sigma^2$ (for fixed $x$) 
by finding the maximizing $\sigma^2$ of~(\ref{eqn:LaplacePrior}).
More specifically, for a given $x$, the maximizing $\sigma^2$ 
of~(\ref{eqn:LaplacePrior}) is
easily determined to be
\begin{IEEEeqnarray}{rCl}
  \hat \sigma^2 &=& \argmax_{\sigma^2} \Normal{x; a, \sigma^2} \rho(\sigma), \\
  &=& \argmin_{\sigma^2} \frac{(x-a)^2}{\sigma^2} + \gamma^2 \sigma^2 \\
  &=& |x-a| / \gamma.
\end{IEEEeqnarray}
Consequently,~(\ref{eqn:LaplacePrior}) amounts to 
\begin{IEEEeqnarray}{rCl} 
  \label{eqn:LaplacePriorExplicit}
  p(x) = \Exp{- \gamma |x-a|},
\end{IEEEeqnarray}
which is proportional to
a Laplace prior, up to a scale factor. 
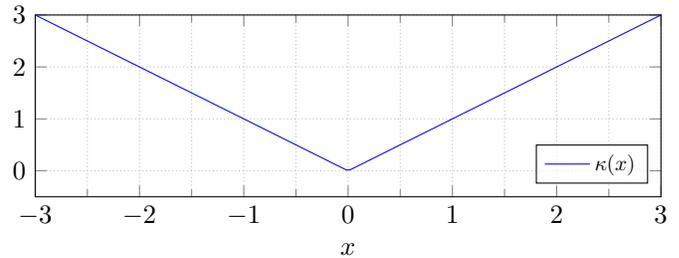
\begin{figure}
\begin{tikzpicture}
\pgfplotsset{minor grid style = {densely dotted}}
\pgfplotsset{major grid style = {densely dotted}}
\begin{axis}[
  width=9.9cm, height=4.cm, 
  xlabel={$x$}, 
  ymin=-0.5,ymax=3, xmin=-3,xmax=3, ylabel near ticks, xlabel near ticks,
  xtick distance = 1, ytick distance =1,
  minor x tick num={1}, 
  grid=both, 
  legend style={at={(0.8,0.18)},anchor=west, fill opacity=0.7, draw
  opacity=1,text opacity=1,nodes={scale=0.8, transform shape}},
]
\addplot[samples=300,color=blue]{1*(abs(x) };
\addlegendentry{$\kappa(x)$}
\end{axis}
\end{tikzpicture}
\caption{The function $\kappa(x) = \gamma |x - a|$, for $\gamma = 1$ and $a =
0$.}
\label{fig:costFuncLaplace}
\end{figure}
The associated cost function is
defined as
\begin{IEEEeqnarray}{rCl}
  \kappa(x) &\eqdef& -\log p(x). \label{eqn:costFuncGen} \\
  &=& \gamma |x-a|, \label{eqn:costFuncLaplace}
\end{IEEEeqnarray}
and is illustrated in Fig.~\ref{fig:costFuncLaplace}.
Throughout the remainder of this document, we will make frequent use of both the
probabilistic view (as in~(\ref{eqn:LaplacePriorExplicit})), and the perspective
of
a cost function (as~(\ref{eqn:costFuncLaplace})).

\section{Box Constraint}
\label{sec:BoxConstraint}

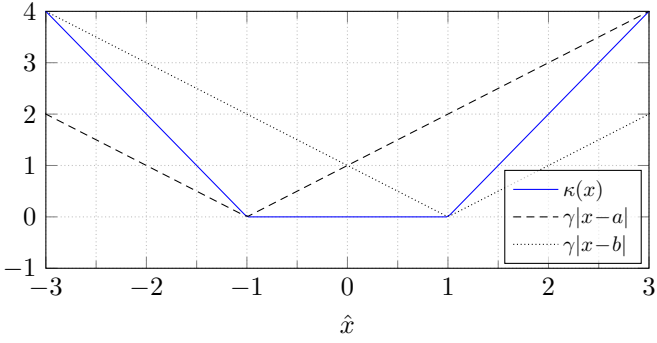
\begin{figure}
\begin{tikzpicture}
\pgfplotsset{minor grid style = {densely dotted}}
\pgfplotsset{major grid style = {densely dotted}}
\pgfplotsset{legend cell align={left}}
\begin{axis}[
  width=9.6cm, height=5.cm, 
  xlabel={$\hat x$}, 
  ymin=-1,ymax=4, xmin=-3,xmax=3, ylabel near ticks, xlabel near ticks,
  xtick distance = 1, ytick distance =1,
  minor x tick num={1}, 
  grid=both, 
  legend style={at={(0.76,0.2)},anchor=west, fill opacity=0.7, draw
  opacity=1,text opacity=1,nodes={scale=0.8, transform shape}},
]
\addplot[samples=300,color=blue]{1*(abs(x+1) + abs(x-1)) - 1*abs(2)};
\addlegendentry{$\kappa( x)$}
\addplot[samples=300,color=black, densely dashed]{1*abs(x+1)};
\addlegendentry{$\gamma | x\!-\!a|$}
\addplot[samples=300,color=black, densely dotted]{1*abs(x-1)};
\addlegendentry{$\gamma | x\!-\!b|$}
\end{axis}
\end{tikzpicture}
\caption{Function $\kappa(x) = \gamma ( |x \!-\! a| + |x \!-\!
b| - |b\!-\!a|)$, for $a=-1, b=1$ and $\gamma = 1$. }
\label{fig:CostBoxPrior}
\end{figure}

The NUV representation of the box constraint is an almost 
obvious combination of two well-known ideas:  
Namely, (i) the NUV representation of the Laplace 
prior of Section~\ref{sec:LaplacePrior},
and (ii) adding two cost functions of the form~(\ref{eqn:costFuncLaplace}) to a
cost function that is constant for $x \in [a, b]$, 
as illustrated in Fig.~\ref{fig:CostBoxPrior}.
In formal terms, we consider a composite prior model of the form
\begin{IEEEeqnarray}{rCl} \label{eqn:BoxPriorExplicit}
  p(x) 
  &\eqdef& \max_\theta \Normal{x; a, \va} \Normal{x; b, \vb}  
    \rho(\sigma_a) \rho(\sigma_b) \psi(\gamma, a, b), \nonumber \\
\end{IEEEeqnarray}
where $\theta = (\va, \vb)$, and where
\begin{IEEEeqnarray}{rCl}
  \rho(\sigma_i) \eqdef \sqrt{2 \pi \sigma_i^2} \exp(- \gamma^2 \sigma_i^2 / 2),
  \label{eqn:BoxAuxFunc}
\end{IEEEeqnarray}
for $i \in \{a,b\}$.
The term $\psi(\gamma, a, b)$ is defined as
\begin{IEEEeqnarray}{rCl}
  \psi(\gamma, a, b) \eqdef \exp(\gamma |b - a|), \label{eqn:CorrectionTerm}
\end{IEEEeqnarray}
for reasons which will be obvious later on.
Note that~(\ref{eqn:BoxPriorExplicit}) is Gaussian in $x$, up to a scale factor,
and can equivalently be written as
\begin{IEEEeqnarray}{rCl} \label{eqn:GaussianSeparated}
  p(x) &=& \max_\theta p(x \cond \theta) g(\theta),
\end{IEEEeqnarray}
where
\begin{IEEEeqnarray}{rCl} \label{eqn:GaussianInX}
  p(x \cond \theta) = \Normal{x; \mt, \vt},
\end{IEEEeqnarray}
with 
\begin{IEEEeqnarray}{rCl}
  \mt &=& \frac{a \vb + b \va}{\va + \vb}, \label{eqn:mTheta}\\
  \vt &=& (1/\va + 1/\vb)^{-1} = \frac{\va \vb}{\va + \vb}, \label{eqn:VTheta}
\end{IEEEeqnarray}
and 
\begin{IEEEeqnarray}{rCl} \label{eqn:ScaleFactor}
  g(\theta) = \Normal{0, a-b, \va+\vb} \rho(\sigma_a) \rho(\sigma_b)
  \psi(\gamma, a, b).
\end{IEEEeqnarray}
Note that $g(\theta)$ is independent of $x$.

For a given $x$, the maximizing $\theta$ of~(\ref{eqn:BoxPriorExplicit}) is
easily determined to be 
\begin{IEEEeqnarray}{rCl} \label{eqn:ThetaHat}
  \hat \theta = (\vah, \vbh) = \big( |x - a|/\gamma, 
  |x - b| / \gamma \big). \label{eqn:ThetaHatBox}
\end{IEEEeqnarray}
Plugging in~(\ref{eqn:ThetaHat}) into~(\ref{eqn:BoxPriorExplicit}) leads to 
\begin{IEEEeqnarray}{rCl}
  p(x) = \Exp{-\gamma(| x-a| + | x-b| - |b - a|)},
\end{IEEEeqnarray}
and consequently, the associated cost function amounts to
\begin{IEEEeqnarray}{rCl}
  \kappa(x) &\eqdef& -\log p(x) = \gamma(| x-a| + | x-b| - |b - a|), 
  \label{eqn:CostFunctionBoxPrior}
\end{IEEEeqnarray}
which is illustrated in Fig.~\ref{fig:CostBoxPrior}. The free parameter
$\gamma$ is used to obtain arbitrarily steep side lobes of $\kappa(x)$,
for $x \notin [a,b]$. The term~(\ref{eqn:CorrectionTerm}) 
shifts~(\ref{eqn:CostFunctionBoxPrior}), such
that $k(x) = 0$, for $x \in [a, b]$. 

\begin{figure}
\centering
\includegraphics[width=0.85\linewidth]{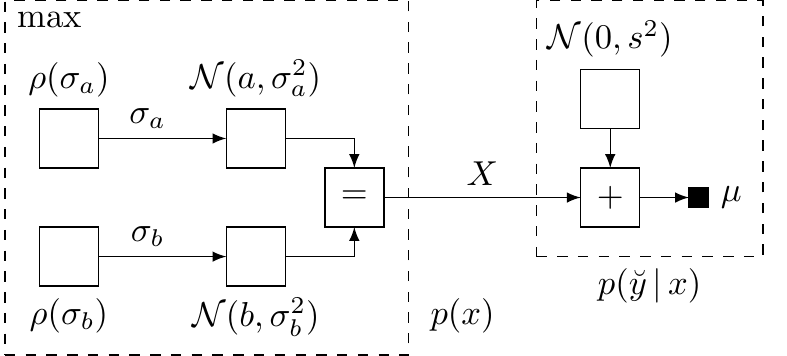}
\caption{Factor graph of the statistical model~(\ref{eqn:JointMAPEstimate}).}
\label{fig:FGTrivialExample}
\end{figure}

We now examine the effect of this proposed prior model by assuming that 
$p(x)$ is used in some simple model with fixed 
observation(s) $\breve y$ and likelihood $p(\breve y \cond x)$, given by 
\begin{IEEEeqnarray}{rCl}  \label{eqn:LLFunction}
 p(\breve y \cond x) = \Normal{x; \mu, s^2}, 
\end{IEEEeqnarray}
where $\mu$ and $s^2 > 0$ depend on $\breve y$.
We assume that $x$ and $\theta$ are determined by joint Maximum-a-Posteriori 
(MAP) estimation according to
\begin{IEEEeqnarray}{rCl} \label{eqn:JointMAPEstimate}
  \hat x  
  &=& \argmax_{x} \max_\theta  p(\breve y \cond x)  p(x \cond \theta) g
  (\theta).
\end{IEEEeqnarray}
The statistical model~(\ref{eqn:JointMAPEstimate}) is 
illustrated as factor graph~\cite{loeliger_introduction_2004} in 
Fig.~\ref{fig:FGTrivialExample}.
The structure of~(\ref{eqn:JointMAPEstimate}) suggests algorithms
that iterate between a maximization step over $x$ for
 fixed
$\theta = \hat \theta$,
and a maximization step over $\theta$ for fixed $x = \hat x$.
The first step is entirely Gaussian, since for fixed $\theta$, 
(\ref{eqn:JointMAPEstimate}) is Gaussian, up to a scale factor.
In the second step, we compute 
\begin{IEEEeqnarray}{rCl}
  \hat \theta = \argmax_\theta p(\breve y \cond x) p(x \cond \theta) g
  (\theta) = \argmax_\theta p(x \cond \theta) g(\theta), \IEEEeqnarraynumspace
\end{IEEEeqnarray}
which coincides with~(\ref{eqn:ThetaHatBox}).

\begin{figure}
\includegraphics{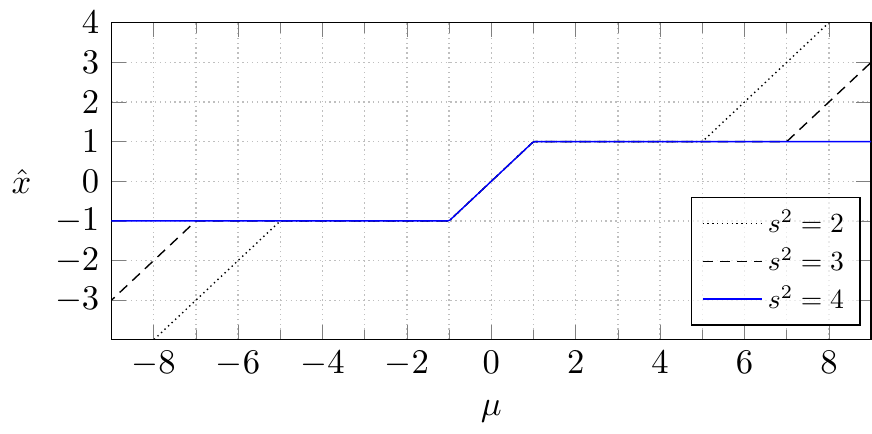}
\caption{Estimate~(\ref{eqn:JointMAPEstimate}) for $a =-1$,
$b=1$, $\gamma=1$, and different $s^2$.}
\label{fig:boxConstChar}
\end{figure}

Note that any such coordinate descent algorithm is guaranteed to converge to a
local maximum or a saddle point, if the underlying objective function is smooth.
Numerical results of~(\ref{eqn:JointMAPEstimate}) are plotted in 
Fig.~\ref{fig:boxConstChar}, for $a=-1$, $b=1$, $\gamma=1$, and
different values for $s^2$.
We observe (and it can be proven) that for a given $\mu$ and $s^2$, and a
sufficiently large $\gamma$, 
the estimate $\hat x$ lies in $[a,b]$. More specifically, the constraint~(\ref{eqn:BoxCon}) 
is satisfied, if and only if
\begin{IEEEeqnarray}{rCl} \label{eqn:HardCondition}
  s^2 \geq  \begin{cases}
    0, & \text{if $\mu \in [a,b]$} \\
    \min \left \{ \frac{|a-\mu|}{2\gamma}, 
                  \frac{|b - \mu|}{2\gamma} \right \}, &
    \mathrm{else}.
    \end{cases} \IEEEeqnarraynumspace
\end{IEEEeqnarray}
Note that since $\gamma$ is a free design parameter,~(\ref{eqn:HardCondition})
can
essentially always be satisfied.


\section{Half-Space Constraint}
\label{sec:HalfSpaceConstraint}
A half-space constraint can be obtained by using a box constraint as in
Section~\ref{sec:BoxConstraint} and letting
one of the boundary points go to $\pm \infty$. We first consider the case for
$\binf$.

For $\binf$,~(\ref{eqn:ThetaHat}) is obviously not well-defined. 
However, if we consider~(\ref{eqn:GaussianSeparated}) with~(\ref{eqn:ThetaHat})
plugged in for $\binf$,~(\ref{eqn:GaussianSeparated}) remains well-defined.
Specifically, we consider a prior of the form
\begin{IEEEeqnarray}{rCl}
  p(x) &\eqdef& \limb \max_{\theta} p(x \cond \theta) g(\theta) 
  \label{eqn:PriorHalfSpace} \\
  &=& \limb  p(x \cond \hat \theta) g(\hat \theta), 
  \label{eqn:LimitDensPluggedIn}
\end{IEEEeqnarray}
where $p(x\cond \theta)$ is as in~(\ref{eqn:GaussianInX}), $g(\theta)$ is 
as in~(\ref{eqn:ScaleFactor}), and $\hat \theta$ as in~(\ref{eqn:ThetaHatBox}).
To show that~(\ref{eqn:PriorHalfSpace}) is well-defined,
we first inspect the first two moments of $p(x \cond \hat
\theta)$ in the limit of $\binf$, i.e.,
\begin{IEEEeqnarray}{rCl}
\mtt &\eqdef& \limb \mt \Big \rvert_{\theta = \hat \theta} \\
&=& \limb \frac{a \vbh + b \vah}{\vah + \vbh} \\
&=& \limb \frac{a |x - b| + b |x - a|}{|x - b| + |x - a|} \\
&=& \limb \frac{ b (|x - a| + a) - a x}{b -
x + |x - a |} \\
&\stackrel{\text{d'H\^{o}pital}}{=}&  |x - a| + a \label{eqn:MeanLimitRight}
\end{IEEEeqnarray}
and
\begin{IEEEeqnarray}{rCl}
  \vtt &=& \limb \vt \Big \rvert_{\theta = \hat \theta} \\
  &=& \limb \frac{\vah \vbh}{\vah + \vbh} \\
  &=& \limb \frac{ |x - a| \cdot |x - b| }
  {\gamma (|x - a| + |x - b|)} \\
  &=& \limb \frac{ |x - a| \cdot (b - x )}
  {\gamma (|x - a| + (b - x) )} \\
  &\stackrel{\text{d'H\^{o}pital}}{=}& |x - a| / \gamma. \label{eqn:VarLimitRight}
\end{IEEEeqnarray}
As a consequence, 
\begin{IEEEeqnarray}{rCl}
  \limb p(x \cond \hat \theta) = \Normal{x; \mtt, \vtt} 
  \label{eqn:GaussianPartLimit}
\end{IEEEeqnarray}
which thus also well-defined.
It remains to prove that $g(\hat \theta)$ in~(\ref{eqn:LimitDensPluggedIn}) 
for $\binf$ is also well-defined. For the sake of simplicity, we first consider 
\begin{IEEEeqnarray}{rll}
  \limb &&-\log g(\hat \theta) \nonumber \\
  &=& \limb
      \frac{1}{2}\log(2 \pi (\vah + \vbh)) + \frac{(a -
      b)^2}{2 (\vah + \vbh)} \IEEEeqnarraynumspace
      \nonumber \\
     && -\frac{1}{2}\log(2 \pi \vah) + \frac{\gamma^2 \vah}{2}
     \nonumber  \\
     && -\frac{1}{2}\log(2 \pi \vbh)  + \frac{\gamma^2 \vbh}{2}
     - \gamma | b - a| \\
  &=& \limb
      \frac{1}{2}\log\left(\frac{\vah}{\vbh} + 1\right) + \frac{(a -
      b)^2}{2 (\vah + \vbh)} \IEEEeqnarraynumspace
      \nonumber \\
     && -\frac{1}{2}\log(2 \pi \vah) + \frac{\gamma^2 (\vah + \vbh)}{2}
     - \gamma | b - a| \\
  &=& \limb
      \underbrace{\frac{1}{2}\log\left(\frac{|x-a|}{|x-b|} + 1\right)}_
      {\rightarrow 0 \; \text{for}\;  \binf} +
      \frac{\gamma (a - b)^2}{2 (|x-a|+|x-b|)}
      \IEEEeqnarraynumspace
      \nonumber \\
     && -\frac{1}{2}\log\left(2 \pi \frac{|x-a|}{\gamma}\right) 
     + \frac{\gamma (|x-a|+|x-b|)}{2} \nonumber\\    
     && - \gamma | b - a|  \\
  &=& \limb
      -\frac{1}{2}\log\left(2 \pi \frac{|x-a|}{\gamma}\right)
      \nonumber \\
      &&
      \underbrace{+ \frac{\gamma (a - b)^2}{2 (|x-a|+|x-b|)}
      + \frac{\gamma (|x-a|+|x-b|)}{2} 
      - \gamma | b - a|}_{\rightarrow 0 \; \text{for}\;  \binf} \nonumber \\ 
      \label{eqn:LimitToZero} \\
  &=& -\log \sqrt{ 2 \pi \frac{|x-a|}{\gamma} }, 
  \label{eqn:gThetaFinal}
\end{IEEEeqnarray}  
where the step from~(\ref{eqn:LimitToZero}) to~(\ref{eqn:gThetaFinal}) 
is justified by
\begin{IEEEeqnarray}{rCl}
  && \limb \frac{\gamma (a - b)^2}{2 (|x-a|+|x-b|)} \nonumber \\
      && \quad \quad \quad \quad + \frac{\gamma (|x-a|+|x-b|)}{2}
      - \gamma | b - a|  \\ 
  &=& \limb \frac{
          \splitfrac{
  b^2 (\gamma + \gamma - 2 \gamma) + b \gamma (-2
  a + 2 (|x - a| - x) 
    }{
  - 2(|x- a| - x - a)) + 
  (\dots)}
  }{2 (|x- a| + (b - x))} \IEEEeqnarraynumspace \\
  &=& \limb \frac{b^2 \cdot 0 + b \cdot 0 + (\dots)}
  {2 (|x- a| + (b - x))} \\
  &=& 0.
\end{IEEEeqnarray} 
Consequently, 
\begin{IEEEeqnarray}{rCl}
  \limb g(\hat \theta) &=& \exp( -\limb -\log g(\hat \theta) ) \\
  &=& \sqrt{2 \pi \frac{|x-a|}{\gamma}},
\end{IEEEeqnarray}
and finally,~(\ref{eqn:PriorHalfSpace}) is well-defined.

The prior~(\ref{eqn:PriorHalfSpace}) can thus be explicitly expressed as
\begin{IEEEeqnarray}{rCl} 
  p(x) &=& \limb  p(x \cond \hat \theta) g(\hat \theta) \\
  &=& \Normal{x; \mtt, \vtt} \sqrt{2 \pi \frac{|x-a|}{\gamma}} \\
  &=& \frac{\sqrt{2 \pi \frac{|x-a|}{\gamma}}}
  {\sqrt{2 \pi \frac{|x-a|}{\gamma}}} 
  \exp\left( - \frac{\gamma (x- a - |x - a|)^2}{2 |x -
  a|} \right) \IEEEeqnarraynumspace \\
  &=& \exp \Big( - \gamma \left(|x- a|- (x- a) \right ) \Big ).
\end{IEEEeqnarray}

The associated cost function is then
\begin{IEEEeqnarray}{rCl}
  \kappa(x) &\eqdef& -\log  p(x) \\
  &=& \gamma \left(|x- a|- (x- a) \right),
\end{IEEEeqnarray}
which is illustrated in Fig.~\ref{fig:CostBoxPriorPlanePlus}.
The free parameter $\gamma$ is again used to obtain an arbitrarily steep side
lobe, see Fig.~\ref{fig:CostBoxPriorPlanePlus}. As can be seen, $k(x) =
0$, for $x \geq a$. For $x < a$, the cost function increases linearly, with a
slope which is determined by $\gamma$.

\begin{figure}
\begin{tikzpicture}
\pgfplotsset{minor grid style = {densely dotted}}
\pgfplotsset{major grid style = {densely dotted}}
\pgfplotsset{legend cell align={left}}
\begin{axis}[
  width=9.6cm, height=4.5cm, 
  xlabel={$ x$}, 
  ymin=-1,ymax=5, xmin=-3,xmax=3, ylabel near ticks, xlabel near ticks,
  xtick distance = 1, ytick distance =1,
  minor x tick num={1}, 
  grid=both, 
  legend pos=north east,
  legend style={fill opacity=0.7, draw
  opacity=1,text opacity=1,nodes={scale=0.8, transform shape}},
]
\addplot[samples=300,color=blue]{1*(abs(x)- x))};
\addlegendentry{$\gamma = 1$}
\addplot[samples=300,color=black, densely dashed]{2*(abs(x)- x))};
\addlegendentry{$\gamma = 2$}
\addplot[samples=300,color=black, densely dotted, line width=0.75pt]{50*(abs
(x)-x))};
\addlegendentry{$\gamma = 50$}
\end{axis}
\end{tikzpicture}
\caption{Function $\kappa(x) = \gamma (|x - a| - (x - a))$, for $a = 0$ 
(right-sided half-plane constraint).
}
\label{fig:CostBoxPriorPlanePlus}
\vskip1em
\begin{tikzpicture}
\pgfplotsset{minor grid style = {densely dotted}}
\pgfplotsset{major grid style = {densely dotted}}
\pgfplotsset{legend cell align={left}}
\begin{axis}[
  width=9.6cm, height=4.5cm, 
  xlabel={$ x$}, 
  ymin=-1,ymax=5, xmin=-3,xmax=3, ylabel near ticks, xlabel near ticks,
  xtick distance = 1, ytick distance =1,
  minor x tick num={1}, 
  grid=both, 
  legend pos=north west,
  legend style={fill opacity=0.7, draw
  opacity=1,text opacity=1,nodes={scale=0.8, transform shape}},
]
\addplot[samples=300,color=blue]{1*(abs(x)+ x))};
\addlegendentry{$\gamma = 1$}
\addplot[samples=300,color=black, densely dashed]{2*(abs(x) + x))};
\addlegendentry{$\gamma = 2$}
\addplot[samples=300,color=black, densely dotted, line width=0.75pt]{50*(abs
(x)+x))};
\addlegendentry{$\gamma = 50$}
\end{axis}
\end{tikzpicture}
\caption{Function $\kappa(x) = \gamma 
(|x - a| + (x - a))$, for $a = 0$ (left-sided half-plane constraint).
}
\label{fig:CostBoxPriorPlaneMinus}
\end{figure}

For a left-sided half-space constraint, i.e., $b \rightarrow -\infty$, 
the derivation can be carried out analogously,
and yields
\begin{IEEEeqnarray}{rCl}
  \mtt \eqdef -| x - a| + a \label{eqn:MeanLimitLeft}
\end{IEEEeqnarray}
and
\begin{IEEEeqnarray}{rCl}
  \vtt \eqdef \frac{ |x - a|}{\gamma}. \label{eqn:VarLimitLeft}
\end{IEEEeqnarray}
The associated cost function amounts to 
\begin{IEEEeqnarray}{rCl}
  \kappa(x) &=& \gamma \left(|x - a| + (x - a) \right), 
\end{IEEEeqnarray}
which is illustrated in Fig.~\ref{fig:CostBoxPriorPlaneMinus}.

The effect of the proposed prior model is again examined by assuming 
that~(\ref{eqn:PriorHalfSpace})
is used in some simple model, analog to the example of the previous section. 
The statistical problem we are solving is the joint MAP estimate of $x$ and
$\theta$, i.e.,
\begin{IEEEeqnarray}{rCl} \label{eqn:HPstatOptProblem}
  \hat x = \argmax_{x} p(\breve y \cond x ) \limb \max_\theta p(x \cond
  \theta) g(\theta), \label{eqn:JointMAPEstimateHP}
\end{IEEEeqnarray}
where $p(\breve y \cond x)$ is as in~(\ref{eqn:LLFunction}). 

\begin{figure}
\includegraphics{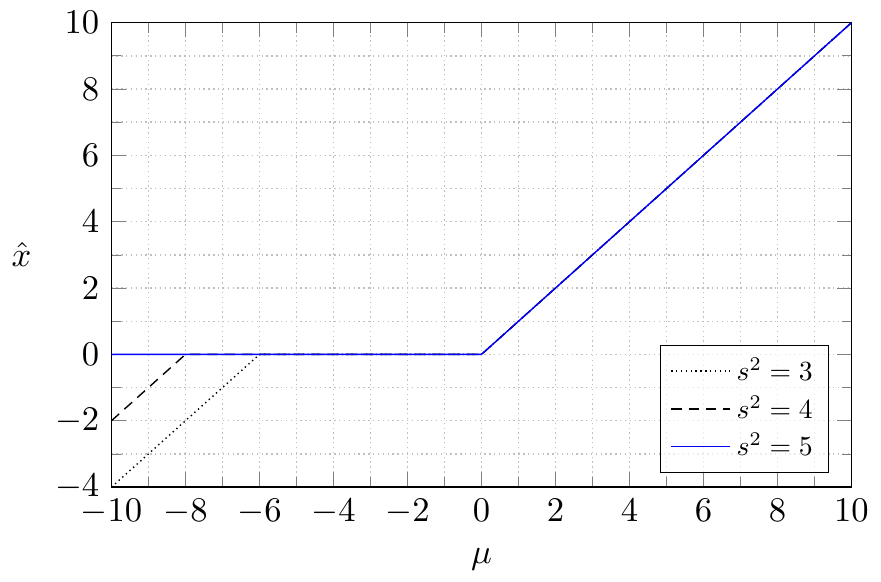}
\caption{Solving~(\ref{eqn:HPstatOptProblem}) numerically for $a=0$ and
$\gamma=1$.}
\label{fig:halfPlaneConstraintChar}
\end{figure}

The structure of~(\ref{eqn:JointMAPEstimateHP}) suggests again algorithms
that iterate between a maximization step over $x$ for
fixed $\theta = \hat \theta$,
and a maximization step over $\theta$ for fixed $x = \hat x$, both in the limit
for $\binf$. In essence, for fixed $x = \hat x$, we directly 
update~(\ref{eqn:GaussianPartLimit}) by computing~(\ref{eqn:MeanLimitRight})
and~(\ref{eqn:VarLimitRight}) (or~(\ref{eqn:MeanLimitLeft}) 
and~(\ref{eqn:VarLimitLeft}), respectively). For fixed
$\theta = \hat \theta$, $g(\theta)$ is an
irrelevant
scale factor and $x$ is determined by
\begin{IEEEeqnarray}{rCl}
  \hat x &=& \argmax_x p(\breve y \cond x) \limb p(x \cond \hat \theta) g
  (\theta) \\
    &=&\argmax_x p(\breve y \cond x) \Normal{x; \mtt, \vtt},
\end{IEEEeqnarray}
which is entirely Gaussian. Numerical results of~(\ref{eqn:JointMAPEstimateHP})
for a right-sided constraint are given in Fig.~\ref{fig:halfPlaneConstraintChar}.

We observe (and it can be proven) that for a given $\mu$ and $s^2$, and a
sufficiently large $\gamma$, 
the estimate $\hat x$ is in $[a, \infty)$. More
specifically, $\hat x \geq a$ is satisfied, if and only if
\begin{IEEEeqnarray}{rCl} \label{eqn:HardConditionHPRight}
  s^2 \geq  \begin{cases}
    0, & \text{if $\mu \geq a$} \\
    \frac{|a-\mu|}{2\gamma}, &
    \mathrm{else}.
    \end{cases} \IEEEeqnarraynumspace
\end{IEEEeqnarray}
Analogously, for a left-sided constraint, we have that 
$\hat x \leq a$ is satisfied, if and only if 
\begin{IEEEeqnarray}{rCl} \label{eqn:HardConditionHPLeft}
  s^2 \geq  \begin{cases}
    0, & \text{if $\mu \leq a$} \\
    \frac{|a-\mu|}{2\gamma}, &
    \mathrm{else}.
    \end{cases} \IEEEeqnarraynumspace
\end{IEEEeqnarray}
Note that since $\gamma$ is a free design 
parameter,~(\ref{eqn:HardConditionHPRight}) (or~(\ref{eqn:HardConditionHPLeft}),
respectively)
can essentially always be satisfied.

\section{Applications}
For the following application examples, we will employ the proposed prior models
in larger Gaussian models, mostly based on an underlying state space model
description of some physical model. Due to the recursive structure of the
used state space models, the estimation of $x$ (or other state quantities) can
efficiently be calculated for example by Gaussian message passing.
For more details, the interested reader is referred
 to~\cite{keusch2021binaryNUV}.

In general, we consider systems which evolve according to
\begin{IEEEeqnarray}{rCl}
x_k  &=& A x_{k-1} + B u_k, \\
y_k &=& C x_k
\end{IEEEeqnarray}
where $k\in \{ 1, 2, \ldots, K\}$, and where $A, B, C,
x_k, u_k$ and $y_k$ have appropriate dimensions.

\subsection{Box Constraints on Inputs}
\begin{figure}
\includegraphics{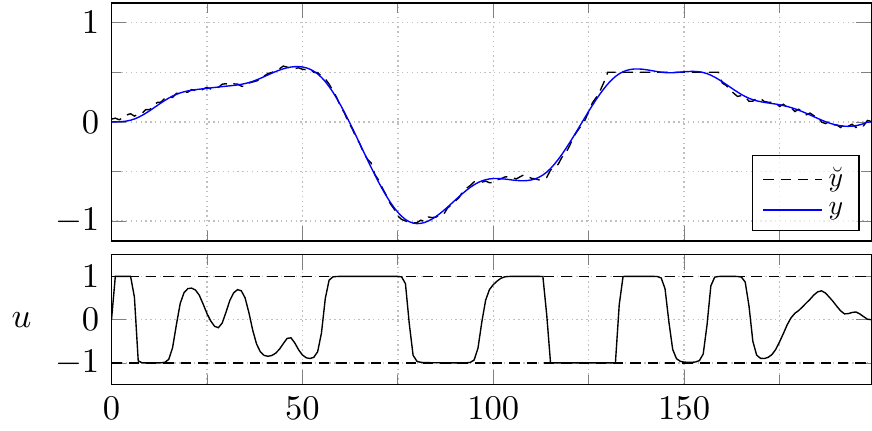}
\caption{Box constraint on input enforcing $u \in [-1, 1]^K$.}
\label{fig:BoxConstraintInput}
\end{figure}

In the example of Fig.~\ref{fig:BoxConstraintInput}, 
the goal is to determine a bounded input sequence $u$ 
such that the system output $y$ approximates a
given target trajectory $\breve y$ (i.e., $\|\breve y - y\|^2$ is minimized).
The underlying linear system is a
third-order low-pass filter (as in~\cite[Section~4.1]{keusch2021binaryNUV}). To achieve this, 
a box constraint is applied on every input
$u_k$, such that $u_k \in [-1, 1]$, for $k \in \{1, \dots, K\}$.

\subsection{Box Constraints on Outputs}
\begin{figure}
\includegraphics{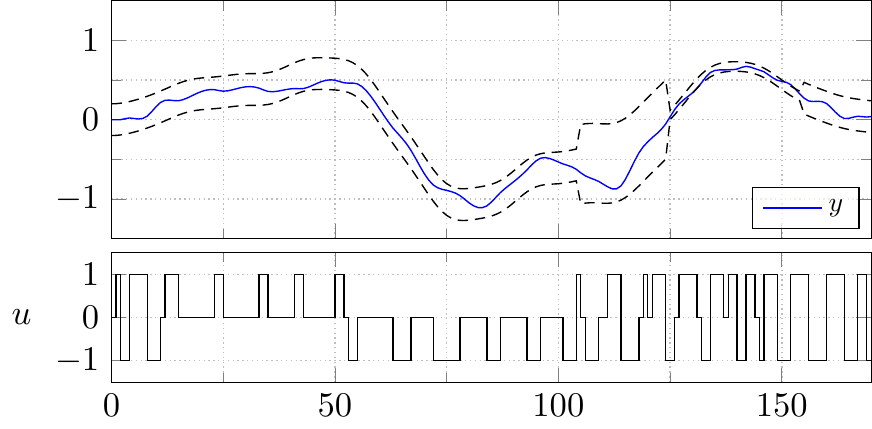}
\caption{Box constraint on output enforcing $y$ to lie in a corridor. Input is $
\{-1,0,1\}$-valued.}
\label{fig:BoxConstraintOutputChannel}
\end{figure}

In the example of Fig.~\ref{fig:BoxConstraintOutputChannel}, 
the goal is to determine a ternary input sequence $u$ (i.e., $u_k \in \{-1, 0,
1\}$) 
such that the system output $y$ lies in a predefined admissible corridor.
The underlying linear system is a
third-order low-pass filter. To achieve this, 
the corridor is modeled by applying box constraints on every output
$y_k$, such that $y_k \in [a_k, b_k]$, for $k \in \{1, \dots, K\}$, 
where $a_k$ and $b_k$ define the corridor at time instance $k$.
The discrete-valued inputs are modeled by binary NUV priors as in~\cite{keusch2021binaryNUV}.

\subsection{Multiple Box Constraints on Outputs I}
\begin{figure}
\includegraphics{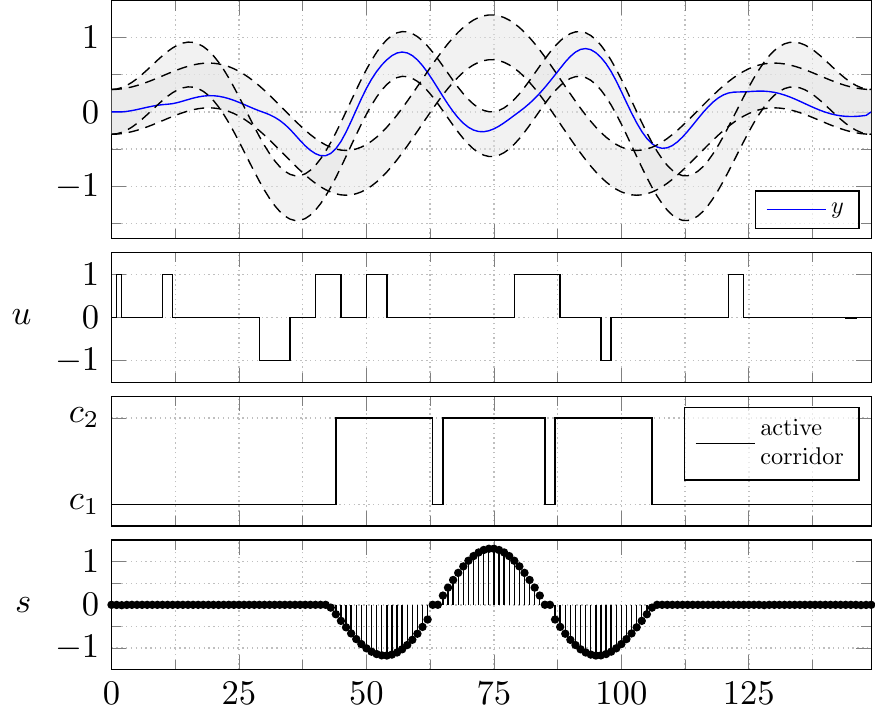}
\caption{Two admissible corridors can be realized by the sum of a binary
decision variable and a box constraint on every output $y_k$. 
The input is $\{-1,0,1\}$-valued. The third plot indicates the ``active'' corridor ($c_1$ or $c_2$).
The last plot illustrates the shift variable $s$. }
\label{fig:boxConstraintOutputMultiple_1}
\end{figure}

\begin{figure}
\centering
\includegraphics[scale=0.85]{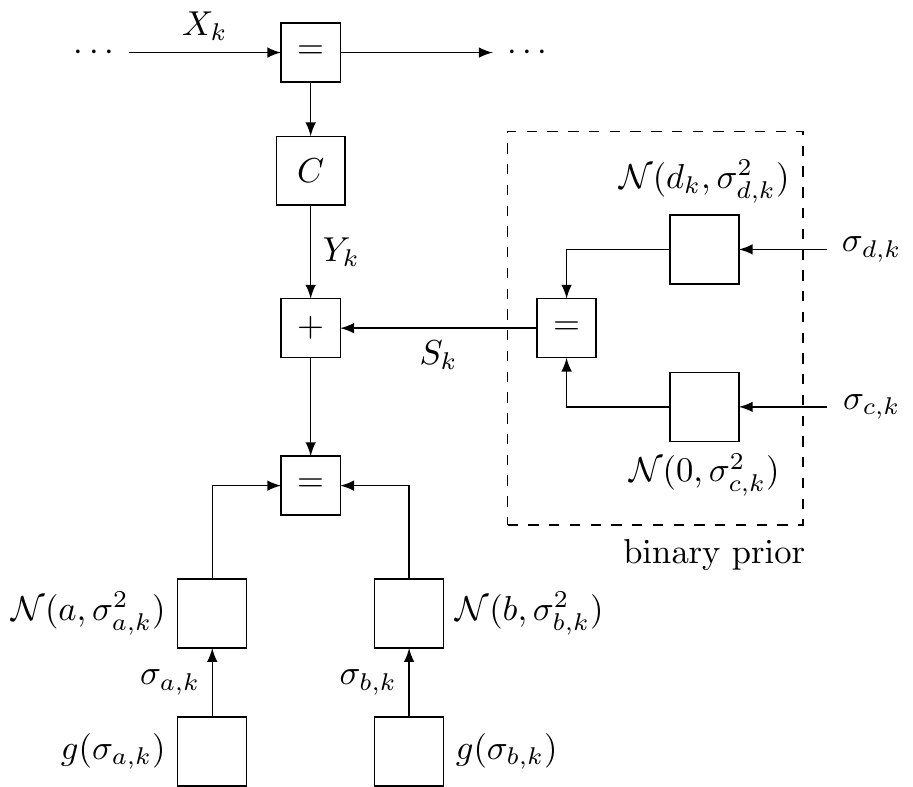}
\caption{Factor graph representing two shifted box constraints.}
\label{fig:fg_multiple_box_constraints}
\end{figure}

In the example of Fig.~\ref{fig:boxConstraintOutputMultiple_1}, 
the goal is to determine a ternary input sequence $u$ (i.e., $u_k \in \{-1, 0,
1\}$) 
such that the system output $y$ lies in either of two predefined admissible
corridors.
If the corridors intersect, the trajectory may switch the corridor, as can be
seen in Fig.~\ref{fig:boxConstraintOutputMultiple_1}.
The underlying linear system is a
third-order low-pass filter. To achieve this, 
the two corridors are modeled as the sum of a binary shift variable 
$S_k \in \{0,d_k\}$ (modeled by a
binary NUV prior~\cite{keusch2021binaryNUV}) and a box constraint, 
such that either $y_k \in [a_k, b_k]$ or $y_k \in [a_k+d_k, b_k+d_k]$, for $k
\in \{1, \dots, K\}$. Note that $d_k$ is in general not constant.
The prior model is given as factor
graph in Fig.~\ref{fig:fg_multiple_box_constraints}.
The discrete-valued inputs are modeled by binary NUV priors as in~\cite{keusch2021binaryNUV}.

\subsection{Multiple Box Constraints on Outputs II (Flappy Bird)}
\begin{figure}
\centering
\includegraphics{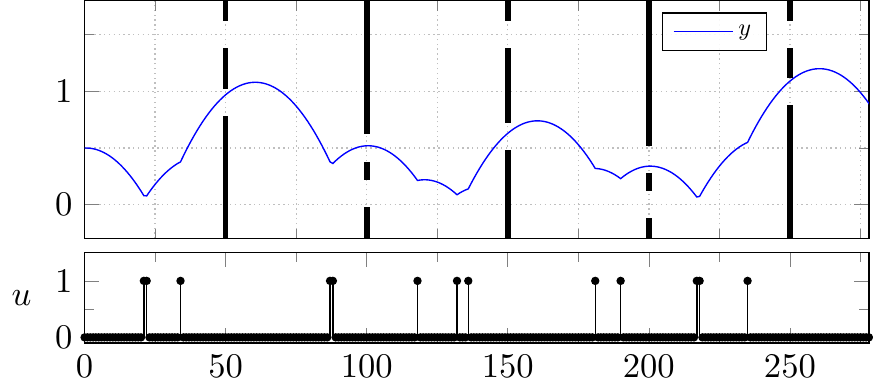}
\caption{Flappy bird control with double-slit obstacles, binary control signal $u$, and resulting trajectory $y$.}
\label{fig:flappy_bird_two_slits}
\end{figure}

In the example of Fig.~\ref{fig:flappy_bird_two_slits}, 
the goal is to ``solve'' a variation of the \emph{flappy bird} 
computer game~\cite{wiki_flappy_bird}.
Consider an analog physical system consisting 
of a point mass $m$ moving forward (left to right in 
Fig.~\ref{fig:flappy_bird_two_slits}) 
with constant horizontal velocity
and ``falling'' vertically with constant acceleration $g$. 
The $\{0,1\}$-valued control signal $u$ affects the system only if $u_k=1$,
in which case a fixed value is added to the vertical momentum.
We wish to steer the point mass such that it passes 
through the double slits, as illustrated in 
Fig.~\ref{fig:flappy_bird_two_slits}. 
To achieve this, the double slits are modeled as in 
Fig.~\ref{fig:fg_multiple_box_constraints}, and the binary 
inputs are modeled by binary NUV priors as in~\cite{keusch2021binaryNUV}.
The underlying dynamical system is given in~\cite[Section~4.2]{keusch2021binaryNUV}.

\subsection{Half-Plane Constraints on Inputs}
\begin{figure}
\includegraphics{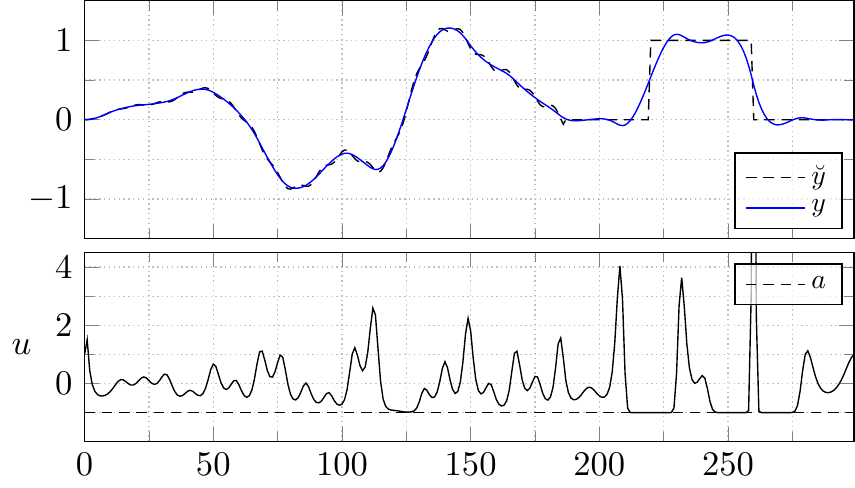}
\caption{Half-space constraint on input enforcing $u_k \geq -1$.}
\label{fig:HalfPlaneConstraintInput}
\end{figure}

In the example of Fig.~\ref{fig:HalfPlaneConstraintInput}, 
the goal is to determine a lower-bounded input sequence $u$ 
such that the system output $y$ approximates a
given target trajectory $\breve y$ (i.e., $\|\breve y - y\|^2$ is minimized).
The underlying linear system is a
third-order low-pass filter. To achieve this, 
a half-plane constraint is applied on every input
$u_k$, such that $u_k \geq -1$, for $k \in \{1, \dots, K\}$.

\subsection{Convex Polyhedrons}

\begin{figure}
\centering
\includegraphics[scale=0.85]{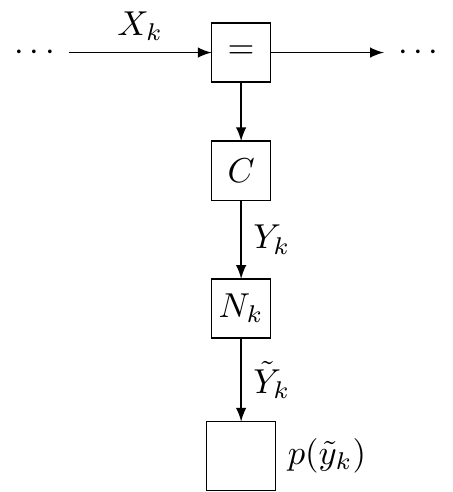}
\caption{Prior model with convex polyhedron constraint, where $\tilde Y_k
\in
\Reals^L$ and where $p(\tilde y_k)$ is the product of $L$ half-space priors.}
\label{fig:FGConvexShapes}
\end{figure}

\begin{figure}
\centering
\includegraphics[scale=0.8]{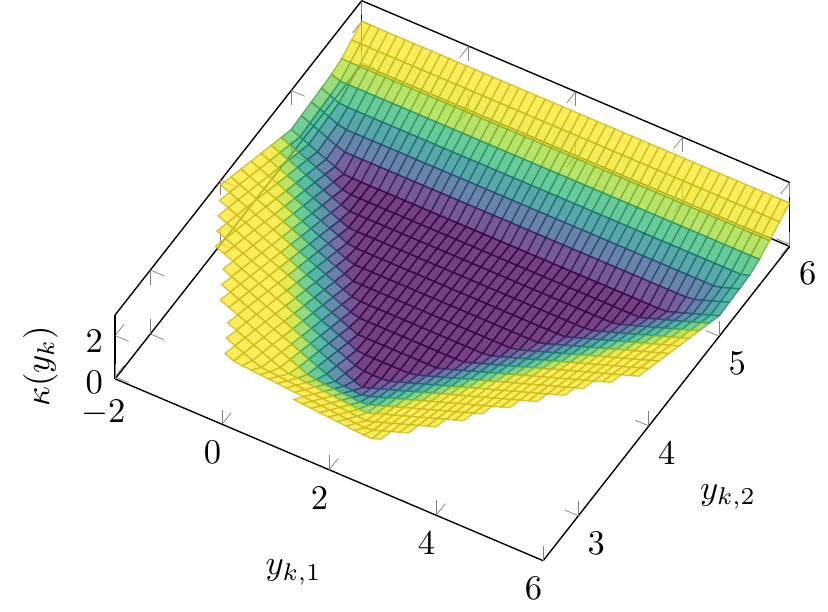}
\caption{Cost function~(\ref{eqn:CostFunction2d}) for $\gamma=1$.}
\label{fig:2dCostFunction}
\end{figure}

\begin{figure}
\centering
\begin{tikzpicture}[scale = 0.7]
   \tkzInit[xmax=5.9,ymax=5.9,xmin=-1.9,ymin=0]
   \begin{scope}[densely dotted]
        \tkzGrid
    \end{scope}
   \tkzAxeXY[/tkzdrawX/label=$y_{k,1}$,/tkzdrawY/label=$y_{k,2}$]

   \draw[thick, -latex] (0, 0) -- (2,3) node[midway,right] {$a_{k,1} n_
   {k,1}$};
   \draw[densely dashed] (-2, 5.666) -- (6,1/3);

   \draw[thick,  -latex] (0, 0) -- (-1,2) node[midway,left] {};
   \node [] at (-1, 0.5) (1) {$a_{k,2} n_{k,2}$};
   \draw[densely dashed] (-2, 1.5) -- (6, 5.5);

   \draw[thick,  -latex] (0, 0) -- (0,5) node[midway,left] {};
   \node [] at (-0.8, 3.5) (1) {$a_{k,3} n_{k,3}$};
   \draw[densely dashed] (-2, 5) -- (6, 5);

   \draw [fill=red, opacity=0.2] (-1, 5) -- (1.575,3.28) -- (5,5) -- cycle;

  \end{tikzpicture}
  \caption{A convex polyhedron (in a two-dimensional setting) defined by three
  normal vectors.}
  \label{fig:ConvexShapeExplainedd}
\end{figure}
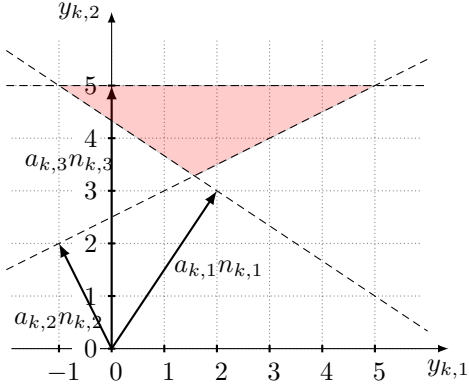

Half-space constraints may be combined to define convex polyhedrons as
admissible regions (see Fig.~\ref{fig:ConvexShapeExplainedd}). 
The factor graph of such an prior model is given in 
Fig.~\ref{fig:FGConvexShapes},
where the constraint is applied at the system output.
The matrix $N_k$ projects the system output $y_k$ onto the normals of $L$
separating hyperplanes and is given by
\begin{IEEEeqnarray}{rCl}
  N_k = \bma n_{k,1} & n_{k, 2} & \cdots & n_{k,L} \ema^\T,
\end{IEEEeqnarray}
where each $n_{k,\ell}$ is a unity-length normal vector. 
The function $p(\tilde y_k)$ is given by
\begin{IEEEeqnarray}{rCl}
  p(\tilde y_k) = \prod_{\ell=1}^L p(\tilde y_{k,\ell}),
\end{IEEEeqnarray}
where each $p(\tilde y_{k,\ell})$ is either a right- or left-sided half-plane
constraint, centered around $a_{\ell}$.
In Fig.~\ref{fig:ConvexShapeExplainedd}
for example, we have $Y_k \in \Reals^2$, $L=3$ and
\begin{IEEEeqnarray}{rClrCl}
  n_{k,1} &=& \bma 2 \\3 \ema / \sqrt{13}, \quad &a_{k,1} &=& \sqrt{13} \\
  n_{k,2} &=& \bma -1 \\ 2 \ema / \sqrt{5}, \quad &a_{k,2} &=& \sqrt{5} \\ 
  n_{k,3} &=& \bma 0 \\ 1 \ema , \quad &a_{k,3} &=& 5,
\end{IEEEeqnarray}
defining a triangle-shaped convex constraint.
The resulting cost function is
\begin{IEEEeqnarray}{rCl} \label{eqn:CostFunction2d}
  \kappa(y_k) &=& 
  \gamma \big(|n_{k,1} y_k - a_{k,1}| - (n_{k,1} y_k - a_
  {k,1}) \nonumber \\
  && +  |n_{k,2} y_k - a_{k,2}| - (n_{k,2} y_k - a_
  {k,2}) \nonumber \\
  && +  |n_{k,3} y_k - a_{k,3}| - (n_{k,3} y_k - a_
  {k,3}) \big),
\end{IEEEeqnarray}
which is illustrated in Fig.~\ref{fig:2dCostFunction}.

\begin{figure}
\includegraphics{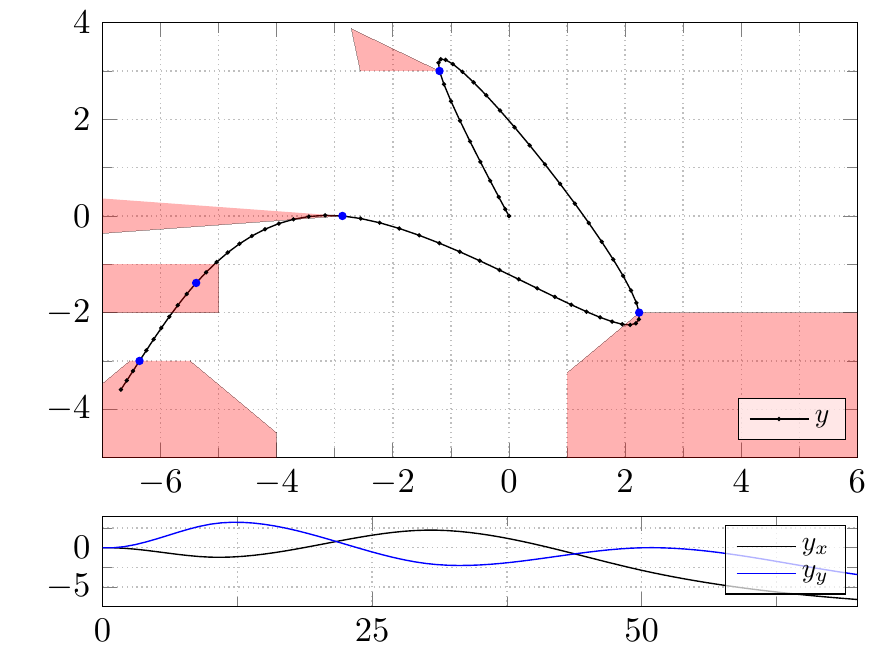}
\caption{Convex polyhedron constraints modeled by multiple half-space
constraints. The constraints enforce the blue points to be within the red
convex polyhedrons.}
\label{fig:HalfPlaneConstraintConvexExample}
\end{figure}

In the example of Fig.~\ref{fig:HalfPlaneConstraintConvexExample}, an object
moves through a two-dimensional space. The goal is that at certain times, the
object's position must lie inside a convex admissible region, where
each region is modeled by a combination of half-plane constraints, as explained
above.

\subsection{Reservoir-Balancing Problem}
\begin{figure}
\centering
\includegraphics[scale=0.9]{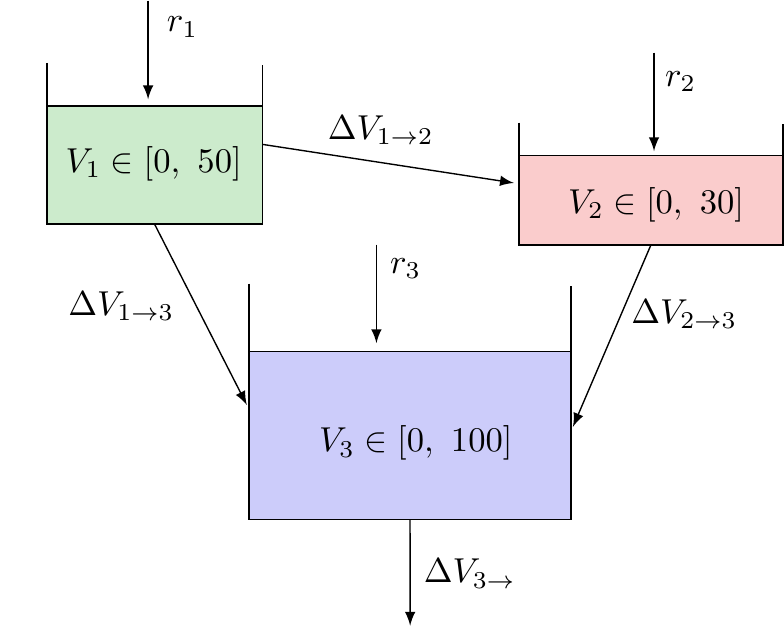}
\caption{Three interconnected water reservoirs with filling levels $V_1, V_2$
and $V_3$. The flow rates between reservoirs are indicated by 
$\Delta V_{(\cdot)}$.}
\label{fig:reservoirProblemDiag}
\end{figure}

\begin{figure}
\includegraphics{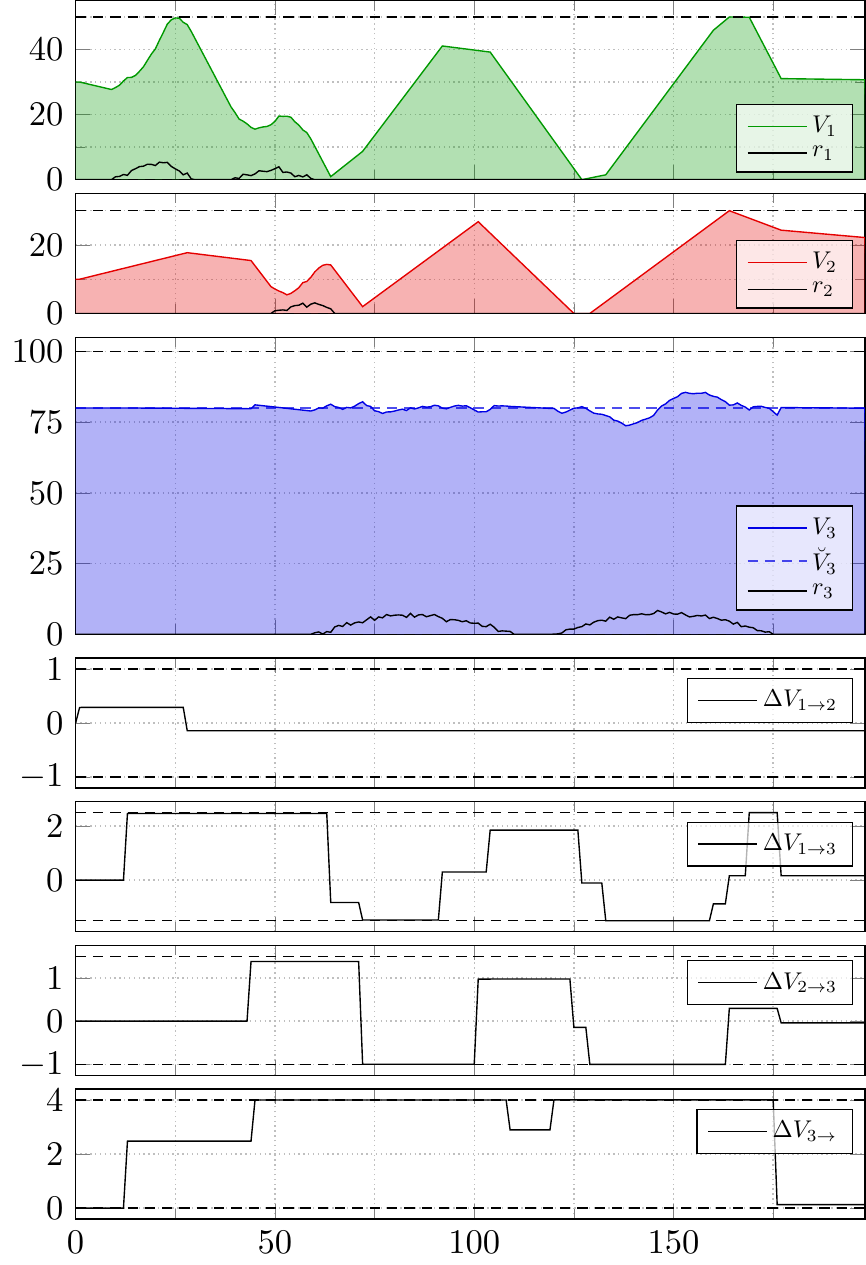}
\caption{Reservoir-balancing problem of three interconnected reservoirs. The
filling levels are indicated by $V_1, V_2$ and $V_3$. Water is pumped between
the reservoirs to minimize $\| V_3 - \breve V_3\|^2$, while keeping the number
of changes in the flow rates sparse.}
\label{fig:reservoirProblem}
\end{figure}

In this example, we consider a system of three interconnected water reservoirs,
as illustrated in Fig.~\ref{fig:reservoirProblemDiag}. Each reservoir has a
maximum filling level ($V_1$, $V_2$ and $V_3$), which must not be exceeded. The
goal of this example is to keep $V_3$ at a constant level of
$\breve V_3 = 80$ (i.e., to minimize $\| V_3 - \breve V_3\|^2$). 
To achieve this, water may be pumped between reservoirs,
where each pump has a maximum achievable flow rate of 
$\Delta V_{1 \rightarrow
2} \in [-1,1]$,
$\Delta V_{1 \rightarrow 3} \in [-1.5,2.5]$,
and
$\Delta V_{2 \rightarrow 3} \in [-1,1.5]$.
In addition, water from $V_3$ may be drained with flow rate $\Delta V_{3
\rightarrow } \in [0,4]$. 
The observable disturbances $r_1, r_2$ and $r_3$ (e.g., rain
forecast) increase the filling levels in $V_1$, $V_2$ and $V_3$, respectively.
The constraints on all filling levels and flow rates are box constraints and
thus easily modeled by the proposed prior of Section~\ref{sec:BoxConstraint}.
Changing the flow rate of a pump (or valve) may be mechanically demanding; thus,
we require changes in the flow rate to occur sparsely, by modeling them
with sparsifying NUV priors~\cite{loeliger_sparsity_2016}.
Numerical results are give in Fig.~\ref{fig:reservoirProblem}. The bounds on
all filling levels and flow rates are indicated by black dashed lines. The
target level $\breve V_3$ is indicated by a blue dashed line in the third plot.
It can be observed that at the time around $k=60$, the disturbance $r_3$
is compensated by draining and pumping water from the third reservoir to the
first and the second. At around $k=140$, a slightly larger disturbance occurs
which cannot be fully compensated by the other reservoirs, which leads to a
significant swing in $V_3$.


\balance

\bibliographystyle{ieeetr}
\bibliography{library}  %
\end{document}